\documentclass[runningheads]{llncs}
\usepackage{verbatim} 
\usepackage{graphicx}
\usepackage{amssymb}
\usepackage{amsmath}
\usepackage{graphicx}

\usepackage{color}
\usepackage{xcolor}
\usepackage[pagebackref=false,breaklinks=true,letterpaper=true,colorlinks,bookmarks=false]{hyperref}

\begin{document}
\title{Domain-specific loss design for unsupervised physical training: A new approach to modeling medical ML solutions}

\titlerunning{Domain-specific loss design for unsupervised physical training}

%
\author{
Hendrik Burwinkel\inst{1}
\and
Holger Matz\inst{4}
\and
Stefan Saur\inst{4}
\and
Christoph Hauger\inst{4}
\and
Ay\c{s}e Mine Evren\inst{1}
\and
Nino Hirnschall\inst{5}
\and
Oliver Findl\inst{5}
\and
Nassir Navab\inst{1,3}
\and
Seyed-Ahmad Ahmadi\inst{2}
}
%
\authorrunning{H. Burwinkel et al.} 
\institute{
Computer Aided Medical Procedures, Technische Universit{\"a}t M{\"u}nchen, Boltzmannstra{\ss}e 3, 85748 Garching bei M{\"u}nchen, Germany\\
\email{hendrik.burwinkel@tum.de}
\and
German Center for Vertigo and Balance Disorders, Ludwig-Maximilians Universit\"at M\"unchen, Marchioninistr. 15, 81377 M\"unchen, Germany
\and
Computer Aided Medical Procedures, Johns Hopkins University, 3400 North Charles Street, Baltimore, MD 21218, USA
\and
Carl Zeiss Meditec AG, Rudolf-Eber-Str. 11, 73447 Oberkochen, Germany
\and
Vienna Hanusch Hospital, Heinrich-Collin-Str. 30, 1140 Vienna, Austria
}

\maketitle              
\begin{abstract}
Today, cataract surgery is the most frequently performed ophthalmic surgery in the world. The cataract, a developing opacity of the human eye lens, constitutes the world's most frequent cause for blindness. During surgery, the lens is removed and replaced by an artificial intraocular lens (IOL). To prevent patients from needing strong visual aids after surgery, a precise prediction of the optical properties of the inserted IOL is crucial. There has been lots of activity towards developing methods to predict these properties from biometric eye data obtained by OCT devices, recently also by employing machine learning. They consider either only biometric data or physical models, but rarely both, and often neglect the IOL geometry. In this work, we propose OpticNet, a novel optical refraction network, loss function, and training scheme which is unsupervised, domain-specific, and physically motivated. We derive a precise light propagation eye model using single-ray raytracing and formulate a differentiable loss function that back-propagates physical gradients into the network. Further, we propose a new transfer learning procedure, which allows unsupervised training on the physical model and fine-tuning of the network on a cohort of real IOL patient cases. We show that our network is not only superior to systems trained with standard procedures but also that our method outperforms the current state of the art in IOL calculation when compared on two biometric data sets.

\end{abstract}
\section{Introduction}
In the field of ophthalmology, the term 'cataract' refers to an internal crystallization of the human eye lens. When untreated, this process develops an increasing and severe opacity of the lens. Due to its ubiquitous appearance especially in older generations, but also for younger populations, it constitutes the most frequent cause for blindness. Fortunately, the treatment of cataracts nowadays is a standard medical procedure and has become the world's most frequently performed ophthalmic surgery. During surgery, the capsular bag of the human eye containing the lens is opened, and the lens is destroyed with ultrasound pulses in a process called phacoemulsification. Then, an artificial intraocular lens (IOL) is inserted into the empty bag, replacing the human lens in the refractive system of the eye. To prevent the patient from the necessity of strong visual aids after the surgery (so called refractive surprises), the optical properties of the IOL have to be defined carefully already prior to surgery. The last years have seen a large body of work to predict these required properties. Today, the prediction is based on biometric data of the patient's eye measured by optical coherence tomography (OCT). Geometric distances like eye length are used to predict an IOL refractive power that satisfies the needs of the eye-specific optical system.\\
These IOL formulas can be clustered into 4 generations. The first and second generation are separated into physical and regression models. The physical models (e.g. Fyodorov, Binkhorst \cite{Binkhorst1979,Fyodorov1975}) are based on the vergence concept, a simple wavefront-based model. The lens is approximated with no thickness and an effective lens position (ELP) \cite{Fyodorov1975}. The second group consists of multilinear regression formulas (SRK I, SRK II \cite{Sanders1980}) based on the measured biometric data. Only the axial eye length (AL) and the keratometry (K), the curvature of the cornea surface, are considered. Still based on the same concepts, in the third generation of IOL formulas (HofferQ, Holladay1, SRK/T \cite{Hoffer1993,Holladay1988,Retzlaff1990}), these basic ideas were further developed by calculating an individual ELP. Additionally, lens constants for the used IOL types were defined. Introduced by formulas like Haigis and Barrett Universal II \cite{Barrett1987,Barrett1993,Turczynowska2016}, in the fourth generation several new biometric measures including the lens thickness (LT) are considered. Still based on the vergence concept, these formulas introduced an increasing amount of parameters to fine-tune the insufficient model. Olsen and Hirnschall et al. tried to overcome these limitations by incorporating raytracing into their models \cite{Hirnschall2019,Olsen2014}. Recently, an increasing number of formulas like the Hill-RBF or Kane also incorporate concepts of machine learning into their predictions \cite{Achiron2017,Clarke1997,Connell2019,Hill2017,Sramka2018,Yarmahmoodi2015}. Due to proprietary interests, most of the details regarding the used methods are not published, however, descriptions of e.g. the Hill-RBF formula on their website describe the model as a big data approach based on over 12.000 patients' eyes \cite{Hill2017}. All current methods are either mainly data-driven or physically motivated models or try to combine both approaches using adjustable parameters or fine-tuning the outcome. Additionally, precise IOL geometry is often neglected.\\
We are overcoming these limitations by transferring precise physical model knowledge into our machine learning system. There has been recent work in the field of physically motivated learning for environment studies and mechanical simulations. In 2017, Karpatne et al. showed that a loss based on physical constraints of temperature and density as well as density and depths can maintain physical consistency and improve the accuracy of the prediction of lake temperature \cite{Karpatne2017}. In 2019, the same group showed an improved RNN model for lake temperature prediction which they initialized by pre-training it on simulated data \cite{Jia2019}. Also using pre-training, Tercan et al. stabilized their prediction of part weights manufactured by injection molding when performed on small datasets \cite{Tercan2018}. Ba et al. \cite{Ba2019} gave an overview of current concepts of physics-based learning, including a work to predict object falls in an image from pure physical constraints \cite{Stewart2017}.\\
\textbf{Contribution.} We propose a new method for physics-based learning in the medical field of ophthalmology for improved  IOL power prediction in cataract surgery. To do so, we derive a detailed single-ray raytracing model of the eye that considers the complete IOL geometry to formulate a domain-specific differentiable loss function which back-propagates physical gradients into our prediction network. This allows an entirely unsupervised training on the physics itself. Further, we propose a transfer learning procedure, which consists of unsupervised training on the physical loss and fine-tuning on real patient cases. We show that the proposed network is not only superior to systems with a standard training procedure but also significantly outperforms the current state of the art in IOL calculation on two biometric data sets. On a wider scope, our work proposes a general methodology for applying medical domain expertise within neural network approaches by designing problem-specific losses. The incorporation of physical models can drastically benefit performance, in particular when only a little amount of supervised training data is available.

\section{Methodology}
\subsection{Mathematical background and unsupervised physical loss}
\textbf{General concept.} Our proposed optical network OpticNet predicts the target value $\textbf{Y} = \text{P}_{\text{IOL}}$, the refractive power of an IOL that leads to a target refraction $\text{Ref}_{\text{T}}$ for an eye with biometric values $\textbf{X}$. It therefore optimises the objective function $f(\textbf{X},\text{Ref}_\text{T}):\textbf{X} \rightarrow \textbf{Y}$. To significantly improve performance, OpticNet explicitly incorporates physical knowledge of the eye's optical system. We design a single-ray raytracer for the calculation of an IOL curvature radius $R$ corresponding to a power $\text{P}_{\text{IOL}}$ minimizing the refractive error for a given $\textbf{X}$. The raytracer is used to design a physical loss $f_{phy}(R,\textbf{X},\text{Ref}_\text{T})$ that backpropagates its physical gradients into the network during an unsupervised training. Essentially, this implements the physical properties of ray propagation in the eye within a neural network. This network is fine-tuned on real-life surgery outcomes from a cohort of patient data to account for deviations from the physical model.
\begin{figure}[t]
    \centering
    \includegraphics[trim = 0px 0px 0px 0px, width=1.0\textwidth]{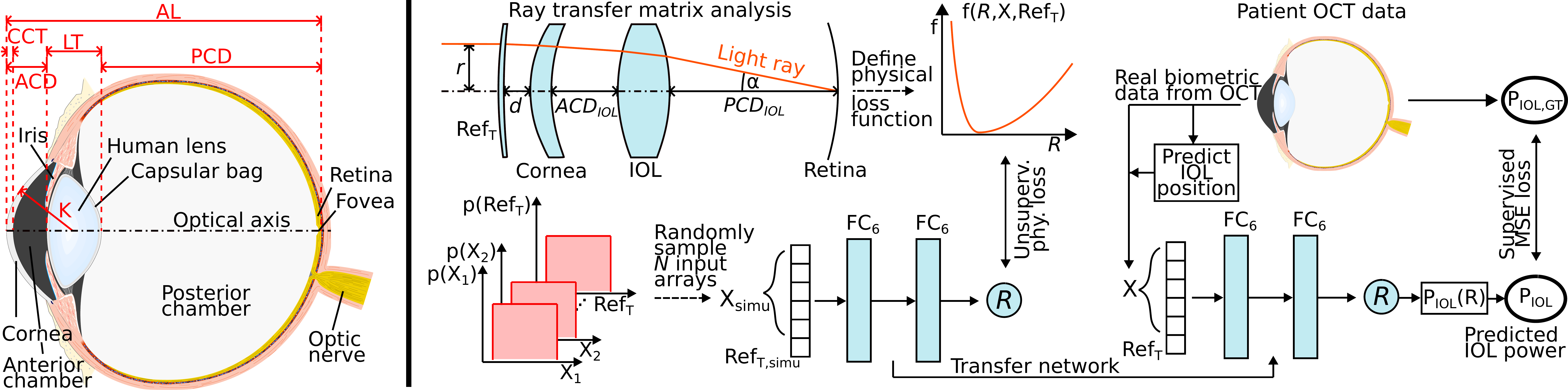}
    \caption{\textbf{Left.} Human eye anatomy. \textbf{Right.} Training: Unsupervised pre-training using physical loss, fine-tuning on real data. AL: axial length, CCT: central cornea thickness, LT: lens thickness, ACD/PCD: anterior/posterior chamber depth, $\text{ACD}_\text{IOL} / \text{PCD}_\text{IOL}$: postoperative ACD/PCD, K: cornea curvature, $\text{FC}_6$: FC layer with 6 neurons.}
    \label{fig:Eye_biometry}
\end{figure}
\textbf{Ray transfer matrix analysis.} In order to describe the physical optical system of the eye, we use the ray transfer matrix analysis. In this physical methodology, every light ray entering an optical system can be described by a vector with two quantities, the distance of its entry point from the optical axis $r$ and its angle $\alpha$ with the axis. Every optical element inside the system is modelled by a matrix either defining the ray propagation within a medium ($A$) or the refraction at a surface between two media ($B$). For an optical path length $x$ inside a medium and a surface with curvature radius $R$ between two media with refractive indices $n_1$ and $n_2$, the corresponding optical elements are modelled by the matrices  \cite{Demtroder2013}:
\begin{equation}
    A = \begin{bmatrix} 1 & x \\ 0 & 1 \end{bmatrix} \hspace{4.0em} B = \begin{bmatrix} 1 & 0 \\ \frac{(\text{n}_\text{1}-\text{n}_\text{2})}{\text{n}_\text{2} \text{R}} & \frac{\text{n}_\text{1}}{\text{n}_\text{2}} \end{bmatrix}
\end{equation}
By successively applying these transformations to the initial ray vector, the propagation of the ray through the system can be formulated (evolution of $r$ and $\alpha$).\\
\textbf{Unsupervised physical loss.} As shown in Fig. \ref{fig:Eye_biometry}, the eye's complex optical system can be divided into the cornea and lens surfaces, as well as their distances inside specific optical media, resulting in the biometric parameters $\textbf{X} = [\text{AL}, ~\text{ACD}_{\text{IOL}}, ~\text{CCT}, ~\text{K}_\text{max}, ~\text{K}_\text{min}]$ and $\text{Ref}_\text{T}$ ($\text{K}_\text{max/min} =$ max. and min. curvature radius of cornea for e.g. astigmatism). We want to achieve that a light ray entering the optical system parallel to the optical axis is focused onto the center of the retina (Fig. \ref{fig:Eye_biometry} right, intersection with retina has distance 0 mm to axis) under an angle $\alpha$, which corresponds to a refractive error of 0 diopter (D). However, often patients have specific wishes for their target refraction $\text{Ref}_\text{T}$ to e.g. read a book easily without any visual aids (e.g. $\text{Ref}_\text{T} = -0.5 ~\text{D}$). To adapt this in the model, an additional very thin refractive element in front of the eye with distance $d$ is used to effectively model this refraction within the system. Due to the thinness, we use the thin lens approximation \cite{Demtroder2013} and formulate it with one matrix as a function of the compensating refraction $-\text{Ref}_\text{T}$. Therefore, the single-ray raytracer is defined as (see Fig. \ref{fig:Eye_biometry}  and its caption for abbreviations):
\begin{multline}
\begin{bmatrix} 0 \\ \alpha \end{bmatrix}
=
\begin{bmatrix} 1 & \text{PCD}_\text{IOL} \\ 0 & 1 \end{bmatrix}
\begin{bmatrix} 1 & 0 \\ \frac{(\text{n}_\text{L}-\text{n}_\text{V})}{-\text{n}_\text{V} R} & \frac{\text{n}_\text{L}}{\text{n}_\text{V}} \end{bmatrix}
\begin{bmatrix} 1 & \text{LT}(R) \\ 0 & 1 \end{bmatrix}
\begin{bmatrix} 1 & 0 \\ \frac{(\text{n}_\text{V}-\text{n}_\text{L})}{\text{n}_\text{L} R} & \frac{\text{n}_\text{V}}{\text{n}_\text{L}} \end{bmatrix}
\begin{bmatrix} 1 & \text{ACD}_\text{IOL} \\ 0 & 1 \end{bmatrix} \cdot \\
\begin{bmatrix} 1 & 0 \\ \frac{(\text{n}_\text{C}-\text{n}_\text{V})}{\text{n}_\text{V} \frac{6.8}{7.7} \text{K}} & \frac{\text{n}_\text{C}}{\text{n}_\text{V}} \end{bmatrix}
\begin{bmatrix} 1 & \text{CCT} \\ 0 & 1 \end{bmatrix}
\begin{bmatrix} 1 & 0 \\ \frac{(1-\text{n}_\text{C})}{\text{n}_\text{C} \text{K}} & \frac{1}{\text{n}_\text{C}} \end{bmatrix}
\begin{bmatrix} 1 & \text{d} \\ 0 & 1 \end{bmatrix}
\begin{bmatrix} 1 & 0 \\ -\text{Ref}_\text{T} & 1 \end{bmatrix}
\cdot
\begin{bmatrix} r \\ 0 \end{bmatrix} =
\textbf{M} \cdot 
\begin{bmatrix} r \\ 0 \end{bmatrix}
\end{multline}
where the IOL thickness LT($R$) is a function of the curvature radius $R$ derived from corresponding lens data, $\text{K} = (\text{K}_\text{max}+\text{K}_\text{min})/2$ is the average cornea curvature, $\text{PCD}_\text{IOL} = \text{AL} - \text{CCT} - \text{ACD}_\text{IOL} - \text{LT}$, $\text{n}_\text{V} = 1.336$ is the refractive index (RI) of the anterior/posterior chamber \cite{Olsen1986}, $\text{n}_\text{C} = 1.376$ is the RI of the cornea \cite{Olsen1986}, $\text{n}_\text{L} = 1.46$ is the RI of the used lens, $\frac{6.8}{7.7}$ is the Gullstrand ratio, the standard ratio of anterior and posterior cornea surface curvature \cite{Olsen1986} and matrix \textbf{M} is the result of the complete matrix multiplication. The refractive power of the IOL is directly described by the curvature radius $R$ inside the equation. After performing the matrix multiplication to receive $\textbf{M}$ and multiplying the ray vector, we obtain two equations. To generate our physical loss, we are only interested in the first one: $0 = \textbf{M}[0,0](R,\textbf{X},\text{Ref}_\text{T}) \cdot r = \textbf{M}[0,0](R,\textbf{X},\text{Ref}_\text{T})$. This equation is linked directly to the IOL power using the optical thick lens formula, which depends on the curvature radius $R$ and yields the corresponding $\text{P}_\text{IOL}$ \cite{Demtroder2013}:
\begin{equation}
    \text{P}_\text{IOL}(R) = (\text{n}_\text{L}-\text{n}_\text{V}) \cdot \left( \frac{2}{R}-\frac{(\text{n}_\text{L}-\text{n}_\text{V}) \cdot \text{LT}(R)}{\text{n}_\text{L} \cdot R^2} \right)
\label{eq:P_calc}
\end{equation}
Therefore, the calculation of $R$ and $\text{P}_\text{IOL}$ are equivalent. We can insert $R$ into Eq. \ref{eq:P_calc} to obtain the searched IOL power. Hence, the function $\textbf{M}[0,0]$ yields zero whenever a radius $R$ and corresponding IOL power is given which results in zero refractive error for an eye with parameters \textbf{X} and $\text{Ref}_\text{T}$. However, a straightforward optimization of a loss defined by the function $\textbf{M}[0,0](R,\textbf{X},\text{Ref}_\text{T})$ would not converge, because it does not have a minimum at its root, but continues into the negative number space. Instead, we use the squared function to keep the loss differentiable and at the same enable minimization using gradient descent to reach the root. Our physical loss is therefore defined by: 
\begin{equation}
    f_{phy}(R,\textbf{X},\text{Ref}_\text{T}) = \textbf{M}[0,0](R,\textbf{X},\text{Ref}_\text{T})^2.
\end{equation}
This physical loss can be fully implemented using differentiable tensors, which allow the propagation of optical refraction gradients into the network. Therefore, it allows an unsupervised training of a network which can predict the physical IOL power for inputs \textbf{X} and $\text{Ref}_\text{T}$. We simply need to sample an arbitrary amount of randomly distributed biometric input data, without any knowledge of the corresponding IOL power ground truth. The network weights are forced to implement the optical refraction, given the loss as a constraint.

\subsection{Training procedure using the unsupervised physical loss}
\textbf{Optical network based on unsupervised physical training.} For every biometric parameter $X_i$ and target refraction $\text{Ref}_\text{T}$, we define a uniform distribution corresponding to a reasonable value range for this parameter. We use these distributions to sample $N$ individual parameter vectors $\textbf{X}_\text{simu}$ and $\text{Ref}_\text{T, simu}$. We properly initialize our network (only positive outcomes to stay in the valid region of $f_{phy}$), train it using these $N$ samples as input and back-propagate the physical loss $f_{phy}(R,\textbf{X},\text{Ref}_\text{T})$ to update the learned weight parameters. A sufficiently large $N$ assures that our network is not overfitting on the data. Nevertheless, we split our $N$ samples into training and testing set to control our training process. The resulting optical network has fully incorporated the physical knowledge and is able to predict the correct physical IOL radius and power for random inputs.\\
\textbf{Training on real data.} Our optical network is fine-tuned on real biometric patient data. A commercial OCT device (ZEISS IOLMaster 700) was used to extract the geometric parameters \textbf{X} for all patient eyes prior and posterior to surgery. Using these as input, the predicted radii $R$ for every patient are processed by Eq. \ref{eq:P_calc} to obtain $\text{P}_\text{IOL}$, then the MSE loss against the IOL power ground truth is backpropagated through Eq. \ref{eq:P_calc} to update the network weights. For every surgery site, the outcome of the surgery due to technique and equipment might slightly differ. Therefore, it is preferable to fine-tune the method on site-specific data. Here, the big advantage of our method comes into play. Instead of handcrafted parameters or training on data neglecting physical knowledge, our optical network has already incorporated this prior knowledge and is able to adapt to the necessary correction as a whole. For a standard untrained network, training on such small amounts of data would inevitably lead to strong overfitting. Due to the raytracing approach, additionally, our loss is capable of incorporating the geometric IOL position $\text{ACD}_\text{IOL}$ instead of a formula-depended ELP. Although unknown prior to surgery, it was shown that the biometric parameters of the eye are correlated to the IOL position \cite{Norrby2017,Hirnschall2018}. The $\text{ACD}_\text{IOL}$ prediction is a research field on its own and not topic of this work. We, therefore, train a PLS model to predict an approximated IOL position for our training using the ground truth of $\text{ACD}_\text{IOL}$, whose usability is an advantage of the method on its own.

\section{Experiments}
\subsection{Experimental setup}
\textbf{Datasets.} For evaluation, we use two biometric datasets obtained during medical studies in two different surgery sites. The datasets have 76 and 130 individual patients correspondingly. The patients did not have any refractive treatment prior to surgery. For every patient, pre- and postoperative OCT images of the patient eye, as well as refractive measurements after surgery, have been collected. We, therefore, have the biometric input, the ground truth for both the position of the IOL and IOL power, as well as the refractive outcome of the surgery.\\
\textbf{Evaluation of methods.} For performance evaluation, we follow the validation method proposed by Wang et al. \cite{Wang2004}. For every patient, the known inserted IOL power and the refractive error of the eye after surgery form a ground truth pair. We are using the measured refractive error as the new target refraction and predict a corresponding IOL power. The difference of the predicted power to the one that was actually inserted in the eye yields the IOL prediction error. We transfer this error to the resulting refractive error using the approximation proposed by Liu et al. \cite{Liu2010}. For every dataset, we perform 10-fold cross-validation. In each run, the data is split into $60\%$ training, $30\%$ validation and $10\%$ test set.\\
\textbf{Network setup.} To correspond to the small input space of six parameters, the neural network is designed as an MLP with 2 hidden layers and 6 hidden units each, mapping down to one output, the predicted IOL power. Settings: learn. rate: $0.001$, w. decay: $0.005$, activation: leakyReLU, $\alpha = 0.1$, optimizer: Adam.

\subsection{Model evaluation}
\begin{figure}[t]
    \centering
    \includegraphics[trim = 0px 0px 0px 0px, width=1.0\textwidth]{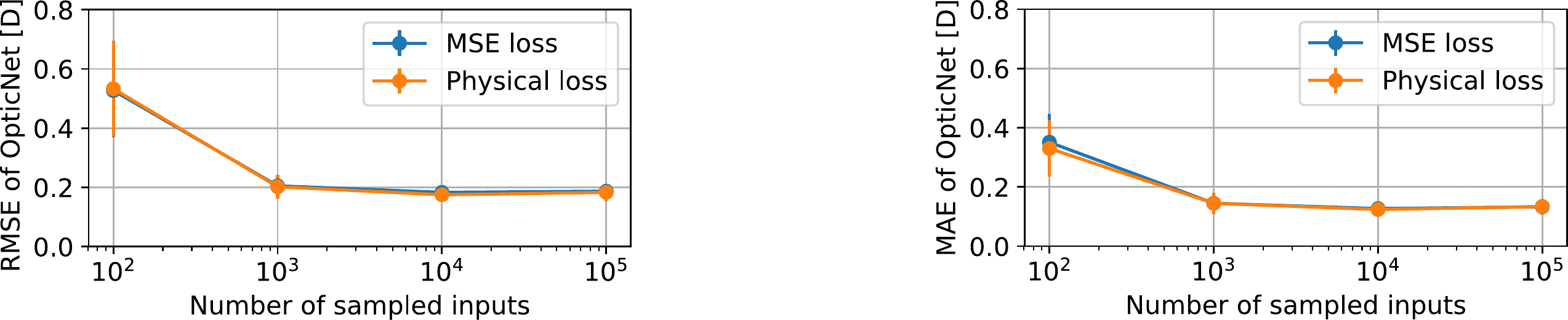}
    \caption{Comparison of physical and MSE loss for different amounts of sampled input.}
    \label{fig:Phy_RMSE}
\end{figure}
\begin{table}[t]
\begin{center}
\caption{Root mean sq. err. and mean abs. err. of IOL prediction and resulting eye refractive error on two biometric datasets (p-value: $\leq$0.05 $\ast$, $<$0.01 $\ast \:  \ast$, $<$0.001 $\ast\ast\ast$).}
\begin{tabular}{ |p{0.8cm}|p{1.7cm}||p{2.0cm}|p{2.0cm}|p{2.0cm}|p{2.0cm}|p{0.7cm}|}
\hline
 Data & Method & RMSE $\text{P}_\text{IOL}$ & MAE $\text{P}_\text{IOL}$ & RMSE Ref. & MAE Ref. & p-val\\
 \hline
 St. 1 & SRK/T & 0.822$\pm$0.256  & 0.692$\pm$0.196 & 0.537$\pm$0.159  & 0.449$\pm$0.124 & $\ast\ast\ast$ \\
 St. 1 & Holliday1 & 0.619$\pm$0.195 & 0.488$\pm$0.138 & 0.405$\pm$0.121 & 0.322$\pm$0.092 & $\ast\:\ast$ \\
 St. 1 & HofferQ & 0.555$\pm$0.149 & 0.416$\pm$0.129 & 0.364$\pm$0.095 & 0.279$\pm$0.084 & \\
 St. 1 & Haigis & 0.538$\pm$0.150  & 0.404$\pm$0.111 & 0.353$\pm$0.092 & 0.267$\pm$0.072 & $\ast$ \\
 St. 1 & Barrett II & 0.556$\pm$0.119  & 0.449$\pm$0.080 & 0.363$\pm$0.073 & 0.295$\pm$0.055 & $\ast\:\ast$ \\
 St. 1 & Raytracer & 0.524$\pm$0.107 & 0.436$\pm$0.087 & 0.338$\pm$0.080 & 0.277$\pm$0.065 & $\ast\:\ast$ \\
 St. 1 & Solo NN & 0.759$\pm$0.650 & 0.595$\pm$0.453 & 0.496$\pm$0.414 & 0.390$\pm$0.293 & $\ast\ast\ast$ \\
 St. 1 & \textbf{OpticNet} & \textbf{0.402}$\pm$\textbf{0.161} & \textbf{0.311}$\pm$\textbf{0.115} & \textbf{0.269}$\pm$\textbf{0.102} & \textbf{0.206}$\pm$\textbf{0.073} & / \\
 \hline
 St. 2 & SRK/T & 0.828$\pm$0.165 & 0.653$\pm$0.123 & 0.540$\pm$0.106 & 0.428$\pm$0.081 & $\ast\ast\ast$ \\
 St. 2 & Holliday1 & 0.746$\pm$0.192 & 0.593$\pm$0.158 & 0.485$\pm$0.126 & 0.388$\pm$0.105 & $\ast\ast\ast$ \\
 St. 2 & HofferQ & 0.758$\pm$0.187 & 0.610$\pm$0.145 & 0.494$\pm$0.122 & 0.399$\pm$0.096 & $\ast\ast\ast$ \\
 St. 2 & Haigis & 0.740$\pm$0.172 & 0.594$\pm$0.132 & 0.482$\pm$0.113 & 0.389$\pm$0.088 & $\ast\ast\ast$ \\
 St. 2 & Barrett II & 0.678$\pm$0.200 & 0.542$\pm$0.162 & 0.441$\pm$0.131 & 0.359$\pm$0.107 & $\ast\:\ast$ \\
 St. 2 & Raytracer & 0.778$\pm$0.126 & 0.616$\pm$0.103 & 0.505$\pm$0.080 & 0.400$\pm$0.063 & $\ast\ast\ast$ \\
 St. 2 & Solo NN & 0.916$\pm$0.578 & 0.630$\pm$0.304 & 0.596$\pm$0.375 & 0.410$\pm$0.200 & $\ast\ast\ast$ \\
 St. 2 & \textbf{OpticNet} & \textbf{0.551}$\pm$\textbf{0.193} & \textbf{0.400}$\pm$\textbf{0.075} & \textbf{0.360}$\pm$\textbf{0.120} & \textbf{0.264}$\pm$\textbf{0.046} & / \\
 \hline
 Both & SRK/T & 0.891$\pm$0.081 & 0.680$\pm$0.080 & 0.581$\pm$0.051 & 0.448$\pm$0.051 & $\ast\ast\ast$ \\
 Both & Holliday1 & 0.834$\pm$0.138 & 0.673$\pm$0.115 & 0.544$\pm$0.082 & 0.439$\pm$0.068 & $\ast\ast\ast$ \\
 Both & HofferQ & 0.820$\pm$0.155 & 0.674$\pm$0.125 & 0.535$\pm$0.093 & 0.44$\pm$0.076 & $\ast\ast\ast$ \\
 Both & Haigis & 0.791$\pm$0.160 & 0.650$\pm$0.130 & 0.516$\pm$0.097 & 0.420$\pm$0.080 & $\ast\ast\ast$ \\
 Both & Barrett II & 0.757$\pm$0.156  & 0.604$\pm$0.129 & 0.494$\pm$0.094 & 0.394$\pm$0.079 & $\ast\ast\ast$ \\
 Both & Raytracer & 0.846$\pm$0.122 & 0.682$\pm$0.102 & 0.553$\pm$0.071 & 0.445$\pm$0.063 & $\ast\ast\ast$ \\
 Both & Solo NN & 0.812$\pm$0.373 & 0.518$\pm$0.199 & 0.529$\pm$0.238 & 0.340$\pm$0.132 & $\ast\ast\ast$ \\
 Both & \textbf{OpticNet} & \textbf{0.562}$\pm$\textbf{0.158} & \textbf{0.376}$\pm$\textbf{0.099} & \textbf{0.366}$\pm$\textbf{0.098} & \textbf{0.249}$\pm$\textbf{0.064} & / \\ 
 \hline
\end{tabular}
\label{tab:IOL_compare}
\end{center}
\end{table}
\textbf{Physical loss against MSE loss.}
We are evaluating the training performance of our unsupervised physical loss against the supervised training with standard MSE loss, where full ground truth has to be present. To obtain this ground truth IOL power for the $N$ randomly generated input samples, we have to solve the physical function $0 = \textbf{M}[0,0](R,\textbf{X},\text{Ref}_\text{T})$ numerically using e.g. the Netwon-Raphson method for every input sample and insert the resulting curvature radius into Eq. \ref{eq:P_calc}. We compare training sizes $N \in \{10^2, 10^3, 10^4, 10^5\}$ 10 times each and use separate validation and test sets of 10000 simulated eyes. Non-converged runs (occurring for $N=10^2$) were discarded (RMSE $>$ 1.0 D). Shown in Fig. \ref{fig:Phy_RMSE}, the physical loss performs equally good or slightly better due to the explicit physical knowledge guiding the training, although it is completely unsupervised and fast to train.\\
\textbf{Performance of OpticNet.} We are comparing the performance of OpticNet against other IOL formulas using our datasets individually and combined. As explained above, unfortunately for many new formulas there is no publication available regarding detailed implementation. Some methods provide online masks for IOL calculation, but due to regulations of the clinical partner usage was not allowed on our patient data. Still, we were able to perform the calculations on several reference IOL formulas, including Barrett Universal II, which is currently considered state-of-the-art, even compared to recent machine learning approaches \cite{Kane2017}. As specified, we perform 10-fold cross-validation, where each test set is unseen prior to its evaluation. For all reference IOL formulas, the latest optimized constants in literature are used. Additionally, we fine-tune every method by calculating the offsets of their averaged predictions on the training set for every fold and subtract this offset from the corresponding test set predictions. Further, we provide the result of our purely physics-based single-ray raytracer (Raytracer) and the performance of a standard network without physical pre-training (Solo NN). As expected and shown in Tab. \ref{tab:IOL_compare}, the Barrett formula has a good performance compared to the other state-of-the-art formulas. However, our OpticNet is significantly outperforming all methods (Wilcoxon signed-rank test). Only for the HofferQ formula on the smaller dataset significance was slightly missed ($\text{p}=0.08$). Especially the NN without pre-training clearly overfits the data, the reported results were only achievable by discarding non-converged runs for the different folds.

\section{Discussion and Conclusion}
In this work, we introduced a new concept for optimized IOL calculation based on the explicit transfer of physical knowledge into our network, called OpticNet. By introducing a new unsupervised physical loss that allows a direct training on the optical model of the eye, a designed single-ray raytracer, physical gradients are propagated into the network. We showed that this physical loss results in equal or improved training performance compared to a standard supervised training, since explicit physical knowledge is incorporated. Additionally, we show that the transfer step on real data significantly outperforms state-of-the-art IOL formulas on two biometric datasets, even stronger when combining the datasets and therefore incorporating the distribution shift. The importance of physical knowledge becomes clear especially in the overfitting of an untrained NN.\\
The concept of unsupervised pre-training on a physical model is a general approach that can be transferred into various fields. It allows the transfer from a solely data-driven training to a model-based training that explicitly incorporates prior knowledge and connects the parameter-based network to the mathematics behind the domain-specific model. This direct connection is ideal for knowledge transfer, since the loss is customized for the task. Shown in this work for IOL calculation, the approach has particular benefits when little annotated training data is available, which is an ubiquitous challenge in the medical domain.\\
\\
\textbf{Acknowledgements}\\
The study was supported by the Carl Zeiss Meditec AG in Oberkochen and Jena, Germany, and the German Federal Ministry of Education and Research (BMBF) in connection with the foundation of the German Center for Vertigo and Balance Disorders (DSGZ) (grant number 01 EO 0901). Further, we thank NVIDIA Corporation for the sponsoring of a Titan V GPU.\\


%
%
%
%
\bibliography{MICCAI2020_ZAI}

\begin{thebibliography}{10}
\providecommand{\url}[1]{\texttt{#1}}
\providecommand{\urlprefix}{URL }
\providecommand{\doi}[1]{https://doi.org/#1}

\bibitem{Achiron2017}
Achiron, A., Gur, Z., Aviv, U., Hilely, A., Mimouni, M., Karmona, L., Rokach,
  L., Kaiserman, I.: {Predicting Refractive Surgery Outcome: Machine Learning
  Approach With Big Data}. Journal of Refractive Surgery  \textbf{33}(9),
  592--597 (sep 2017). \doi{10.3928/1081597X-20170616-03}

\bibitem{Ba2019}
Ba, Y., Zhao, G., Kadambi, A.: {Blending Diverse Physical Priors with Neural
  Networks} (1),  1--15 (2019), \url{http://arxiv.org/abs/1910.00201}

\bibitem{Barrett1987}
Barrett, G.D.: {Intraocular lens calculation formulas for new intraocular lens
  implants}. Journal of Cataract and Refractive Surgery  \textbf{13}(4),
  389--396 (1987). \doi{10.1016/S0886-3350(87)80037-8}

\bibitem{Barrett1993}
Barrett, G.D.: {An improved universal theoretical formula for intraocular lens
  power prediction}. Journal of Cataract and Refractive Surgery
  \textbf{19}(6),  713--720 (1993). \doi{10.1016/S0886-3350(13)80339-2}

\bibitem{Binkhorst1979}
Binkhorst, R.D.: {Intraocular Lens Power Calculation}. International
  Ophthalmology Clinics  \textbf{19}(4),  237--254 (1979).
  \doi{10.1097/00004397-197901940-00010}

\bibitem{Clarke1997}
Clarke, G.P., Burmeister, J.: {Comparison of intraocular lens computations
  using a neural network versus the Holladay formula}. Journal of Cataract and
  Refractive Surgery  \textbf{23}(10),  1585--1589 (1997).
  \doi{10.1016/S0886-3350(97)80034-X}

\bibitem{Connell2019}
Connell, B.J., Kane, J.X.: {Comparison of the Kane formula with existing
  formulas for intraocular lens power selection}. BMJ Open Ophthalmology
  \textbf{4}(1), ~1--6 (2019). \doi{10.1136/bmjophth-2018-000251}

\bibitem{Demtroder2013}
Demtr{\"{o}}der, W.: {Experimentalphysik 2}. Springer-Lehrbuch, Springer Berlin
  Heidelberg, Berlin, Heidelberg (2013). \doi{10.1007/978-3-642-29944-5}

\bibitem{Fyodorov1975}
Fyodorov, S.N., Galin, M.A., Linksz, A.: {Calculation of the optical power of
  intraocular lenses}. Investigative Ophthalmology {\&} Visual Science August
  \textbf{14}(8),  625--628 (1975)

\bibitem{Hill2017}
Hill, W.E.: {Hill-RBF Method}. Haag-Streit White Paper  (2017)

\bibitem{Hirnschall2019}
Hirnschall, N., Buehren, T., Trost, M., Findl, O.: {Pilot evaluation of
  refractive prediction errors associated with a new method for
  ray-tracing–based intraocular lens power calculation}. Journal of Cartaract
  {\&} Refractive Surgery  \textbf{45}(6),  738--744 (2019).
  \doi{10.1016/j.jcrs.2019.01.023}

\bibitem{Hirnschall2018}
Hirnschall, N., Farrokhi, S., Amir-Asgari, S., Hienert, J., Findl, O.:
  {Intraoperative optical coherence tomography measurements of aphakic eyes to
  predict postoperative position of 2 intraocular lens designs}. Journal of
  Cataract and Refractive Surgery  \textbf{44}(11),  1310--1316 (2018).
  \doi{10.1016/j.jcrs.2018.07.044}

\bibitem{Hoffer1993}
Hoffer, K.J.: {The Hoffer Q formula: A comparison of theoretic and regression
  formulas}. Journal of Cataract and Refractive Surgery  \textbf{19}(6),
  700--712 (1993). \doi{10.1016/S0886-3350(13)80338-0}

\bibitem{Holladay1988}
Holladay, J.T., Musgrove, K.H., Prager, T.C., Lewis, J.W., Chandler, T.Y.,
  Ruiz, R.S.: {A three-part system for refining intraocular lens power
  calculations}. Journal of Cataract and Refractive Surgery  \textbf{14}(1),
  17--24 (1988). \doi{10.1016/S0886-3350(88)80059-2}

\bibitem{Jia2019}
Jia, X., Willard, J., Karpatne, A., Read, J., Zwart, J., Steinbach, M., Kumar,
  V.: {Physics guided RNNs for modeling dynamical systems: A case study in
  simulating lake temperature profiles}. SIAM International Conference on Data
  Mining, SDM 2019 pp. 558--566 (2019). \doi{10.1137/1.9781611975673.63}

\bibitem{Kane2017}
Kane, J.X., {Van Heerden}, A., Atik, A., Petsoglou, C.: {Accuracy of 3 new
  methods for intraocular lens power selection.} Journal of cataract and
  refractive surgery  \textbf{43}(3),  333--339 (2017).
  \doi{10.1016/j.jcrs.2016.12.021}

\bibitem{Karpatne2017}
Karpatne, A., Watkins, W., Read, J., Kumar, V.: {Physics-guided Neural Networks
  (PGNN): An Application in Lake Temperature Modeling}  (2017),
  \url{http://arxiv.org/abs/1710.11431}

\bibitem{Liu2010}
Liu, Y., Wang, Z., Mu, G.: {Effects of measurement errors on refractive
  outcomes for pseudophakic eye based on eye model}. Optik  \textbf{121}(15),
  1347--1354 (2010). \doi{10.1016/j.ijleo.2009.01.022}

\bibitem{Norrby2017}
Norrby, S., Bergman, R., Hirnschall, N., Nishi, Y., Findl, O.: {Prediction of
  the true IOL position}. British Journal of Ophthalmology  \textbf{101}(10),
  1440--1446 (2017). \doi{10.1136/bjophthalmol-2016-309543}

\bibitem{Olsen1986}
Olsen, T.: {On the calculation of power from curvature of the cornea}. Br J
  Ophthalmol  \textbf{70}(2),  152--154 (1986). \doi{10.1136/bjo.70.2.152}

\bibitem{Olsen2014}
Olsen, T., Hoffmann, P.: {C constant: New concept for ray tracing-assisted
  intraocular lens power calculation}. Journal of Cataract and Refractive
  Surgery  \textbf{40}(5),  764--773 (2014). \doi{10.1016/j.jcrs.2013.10.037}

\bibitem{Retzlaff1990}
Retzlaff, J.A., Sanders, D.R., Kraff, M.C.: {Development of the SRK/T
  intraocular lens implant power calculation formula}. Journal of Cataract and
  Refractive Surgery  \textbf{16}(3),  333--340 (1990).
  \doi{10.1016/S0886-3350(13)80705-5}

\bibitem{Sanders1980}
Sanders, D.R., Kraff, M.C.: {Improvement of intraocular lens power calculation
  using empirical data}. American Intra-Ocular Implant Society Journal
  \textbf{6}(3),  263--267 (1980). \doi{10.1016/S0146-2776(80)80075-9}

\bibitem{Sramka2018}
Sramka, M., Vlachynska, A.: {Arti?cial Neural Networks Application in
  Intraocular Lens Power Calculation}. In: Proceedings of The 9th EUROSIM
  Congress on Modelling and Simulation, EUROSIM 2016, The 57th SIMS Conference
  on Simulation and Modelling SIMS 2016. vol.~142, pp. 25--30 (dec 2018).
  \doi{10.3384/ecp1714225}

\bibitem{Stewart2017}
Stewart, R., Ermon, S.: {Label-free supervision of neural networks with physics
  and domain knowledge}. 31st AAAI Conference on Artificial Intelligence, AAAI
  2017  \textbf{1}(1),  2576--2582 (2017)

\bibitem{Tercan2018}
Tercan, H., Guajardo, A., Heinisch, J., Thiele, T., Hopmann, C., Meisen, T.:
  {Transfer-Learning: Bridging the Gap between Real and Simulation Data for
  Machine Learning in Injection Molding}. Procedia CIRP  \textbf{72},  185--190
  (2018). \doi{10.1016/j.procir.2018.03.087}

\bibitem{Turczynowska2016}
Turczynowska, M., Ko{\'{z}}lik-Nowakowska, K., Gaca-Wysocka, M., Grzybowski,
  A.: {Effective Ocular Biometry and Intraocular Lens Power Calculation}.
  European Ophthalmic Review  \textbf{10}(02), ~94 (2016).
  \doi{10.17925/EOR.2016.10.02.94}

\bibitem{Wang2004}
Wang, L., Booth, M.a., Koch, D.D.: {Comparison of intraocular lens power
  calculation methods in eyes that have undergone laser-assisted in-situ
  keratomileusis.} Transactions of the American Ophthalmological Society
  \textbf{102},  189--96; discussion 196--7 (2004)

\bibitem{Yarmahmoodi2015}
Yarmahmoodi, M., Arabalibeik, H., Mokhtaran, M., Shojaei, A.: {Intraocular Lens
  Power Formula Selection Using Support Vector Machines}. Frontiers in
  Biomedical Technologies  \textbf{2}(1),  36--44 (2015)

\end{thebibliography}
\end{document}